# Performance Metric for Multiple Anomaly Score Distributions with Discrete Severity Levels

Wonjun Yi
Department of Electrical Engineering
Korea Advanced Institute of Science and Technology
Daejeon, South Korea
lasscap@kaist.ac.kr

Wonho Jung
Department of Mechanical Engineering
Korea Advanced Institute of Science and Technology
Daejeon, South Korea
wonho1456@kaist.ac.kr

Yong-Hwa Park
Department of Mechanical Engineering
Korea Advanced Institute of Science and Technology
Daejeon, South Korea
yhpark@kaist.ac.kr

*Abstract*— The rise of smart factories has heightened the demand for automated maintenance, and normal-data-based anomaly detection has proved particularly effective in environments where anomaly data are scarce. This method, which does not require anomaly data during training, has prompted researchers to focus not only on detecting anomalies but also on classifying severity levels by using anomaly scores. However, the existing performance metrics, such as the area under the receiver operating characteristic curve (AUROC), do not effectively reflect the performance of models in classifying severity levels based on anomaly scores. To address this limitation, we propose the weighted sum of the area under the receiver operating characteristic curve (WS-AUROC), which combines AUROC with a penalty for severity level differences. We conducted various experiments using different penalty assignment methods: uniform penalty regardless of severity level differences, penalty based on severity level index differences, and penalty based on actual physical quantities that cause anomalies. The latter method was the most sensitive. Additionally, we propose an anomaly detector that achieves clear separation of distributions and outperforms the ablation models on the WS-AUROC and AUROC metrics.

## I. Introduction

With the proliferation of smart factories, the demand for automated maintenance has increased significantly. Normal-data-based anomaly detection is effective in environments where anomaly data are scarce because it does not require such data for training and validation. Recently, this approach has evolved not only to detect anomalies but also classify their severity levels based on anomaly scores [1], [2]. Accordingly, systems with higher anomaly scores indicate greater severity levels.

One method to measure performance is by setting a threshold for anomaly scores to classify anomalies by severity level, and then, as in [3], applying penalties for discrepancies between the ground truth and predicted severity levels. This approach assumes that thresholding is a valid method for delineating severity levels. However, setting thresholds inherently involves referencing test data's anomaly scores, which contradicts the goal of avoiding reliance on test data for performance evaluation.

To address these issues, we propose the weighted sum of the area under the receiver operating characteristic curve (WS-AUROC). This method integrates penalties based on severity level differences between two distributions under evaluation with AUROC. We demonstrate that WS-AUROC effectively separates anomaly scores depending on severity levels and is considerably more sensitive than AUROC. Additionally, we compare different penalty assignment methods: uniform penalty regardless of severity level differences, penalty based on severity level index differences, and penalties based on actual physical quantities that cause anomalies. The results indicate that the method based on actual physical quantities is the most sensitive in practice.

To demonstrate the application of WS-AUROC and AUROC, we propose an anomaly detection model that performs well on a vibration dataset [4], which simulates anomalies in a test bed. In this model, the final anomaly score is calculated by summing the anomaly scores obtained through application of the K-nearest neighbor (KNN) method to rotation frequency harmonics and ball pass frequency outer race (BPFO) frequency harmonics and those obtained by inputting fast Fourier transform (FFT) features into a one-dimensional convolutional neural network (1DCNN). According to this model, the anomaly score increases in proportion to the severity level. Additionally, in ablation studies, the proposed anomaly detector outperforms other methods on the AUROC and WS-AUROC metrics. The code is available in [5].

## II. Proposed Metric

The WS-AUROC metric assigns penalties based on the discrepancy between two distributions of anomaly scores corresponding to different severity levels. The rationale is that the greater the discrepancy, the higher should be the penalty.

Let $A$ represent the WS-AUROC metric. For the anomaly severity level index $i \in \{0,1,...,n\}$, where 0 denotes the normal condition, and numbers greater than 0 denote the severity level of an anomaly condition, the anomaly score distribution $S_i$ can be measured. Next, the AUROC $a_{ij}$ between two anomaly score distributions $S_i$ and $S_j$ can be measured. Then, by using multiple $a_{ij}$, WS-AUROC can be calculated as follows:

$$A = 1 - \sum_{i=0}^{n-1} \sum_{j=i+1}^{n} p_{ij}(1 - a_{ij}) = \sum_{i=0}^{n-1} \sum_{j=i+1}^{n} p_{ij} a_{ij} \quad (1)$$

where $p_{ij}$ is the penalty assigned based on the severity level difference between $S_i$ and $S_j$.

TABLE I. PERFORMANCE METRICS (UNBALANCE, BPFI)

| Anomaly type | | Unbalance | | | | BPFI | | | |
|---|---|---|---|---|---|---|---|---|---|
| Index | Combination | AUROC | WS-AUROC (U) | WS-AUROC (I) | WS-AUROC (P) | AUROC | WS-AUROC (U) | WS-AUROC (I) | WS-AUROC (P) |
| 1 | Rotation | 1.000 | 0.992 | 0.997 | 0.997 | 1.000 | 0.502 | 0.601 | 0.445 |
| 2 | BPFI | 0.509 | 0.484 | 0.479 | 0.480 | 0.986 | 0.972 | 0.983 | 0.976 |
| 3 | BPFO | 0.434 | 0.450 | 0.436 | 0.429 | 1.000 | 0.822 | 0.886 | 0.774 |
| 4 | Rotation, BPFO | 0.998 | 0.992 | 0.996 | 0.997 | 1.000 | 0.752 | 0.802 | 0.658 |
| 5 | DL | 0.527 ±0.040 | 0.534 ±0.029 | 0.543 ±0.037 | 0.537 ±0.036 | 1.000 ±0.000 | 0.999 ±0.002 | 0.999 ±0.001 | 1.000 ±0.001 |
| 6 | Rotation, BPFO, DL | 0.997 ±0.002 | 0.993 ±0.003 | 0.997 ±0.001 | 0.998 ±0.001 | 1.000 ±0.000 | 0.994 ±0.005 | 0.996 ±0.003 | 0.992 ±0.006 |

TABLE III. PERFORMANCE METRICS (BPFO, MISALIGNMENT)

| Anomaly type | | BPFO | | | | Misalignment | | | |
|---|---|---|---|---|---|---|---|---|---|
| Index | Combination | AUROC | WS-AUROC (U) | WS-AUROC (I) | WS-AUROC (P) | AUROC | WS-AUROC (U) | WS-AUROC (I) | WS-AUROC (P) |
| 1 | Rotation | 1.000 | 0.979 | 0.988 | 0.974 | 1.000 | 1.000 | 1.000 | 1.000 |
| 2 | BPFI | 1.000 | 0.623 | 0.697 | 0.612 | 0.636 | 0.570 | 0.586 | 0.576 |
| 3 | BPFO | 1.000 | 0.888 | 0.913 | 0.847 | 0.848 | 0.763 | 0.803 | 0.795 |
| 4 | Rotation, BPFO | 1.000 | 0.952 | 0.971 | 0.941 | 1.000 | 1.000 | 1.000 | 1.000 |
| 5 | DL | 1.000 ±0.000 | 1.000 ±0.000 | 1.000 ±0.000 | 1.000 ±0.000 | 0.997 ±0.016 | 0.796 ±0.055 | 0.859 ±0.055 | 0.819 ±0.061 |
| 6 | Rotation, BPFO, DL | 1.000 ±0.000 | 1.000 ±0.000 | 1.000 ±0.000 | 1.000 ±0.000 | 1.000 ±0.000 | 1.000 ±0.000 | 1.000 ±0.000 | 1.000 ±0.000 |

TABLE II. PERFORMANCE METRICS (AVERAGE)

| Anomaly type | | Average | | | |
|---|---|---|---|---|---|
| Index | Combination | AUROC | WS-AUROC (U) | WS-AUROC (I) | WS-AUROC (P) |
| 1 | Rotation | 1.000 | 0.868 | 0.897 | 0.854 |
| 2 | BPFI | 0.783 | 0.662 | 0.686 | 0.661 |
| 3 | BPFO | 0.821 | 0.731 | 0.760 | 0.711 |
| 4 | Rotation, BPFO | 1.000 | 0.924 | 0.942 | 0.899 |
| 5 | DL | 0.881 | 0.832 | 0.850 | 0.839 |
| 6 | Rotation, BPFO, DL | 0.999 | 0.997 | 0.998 | 0.998 |

$p_{ij}$ can be defined in three different ways: uniform (U), index (I), and physics (P). First, a uniform penalty can be applied regardless of severity level differences, as follows:

$$p_{ij} = \frac{2}{n(n+1)}. \quad (2)$$

Second, index penalty can be defined in proportion to the severity level index difference $j-i$, which ensures that larger differences in severity levels lead to higher penalties:

$$p_{ij} = \frac{6(j-i)}{n(n+1)(n+2)}. \quad (3)$$

Third, physics penalty can be based on actual physical quantities that cause anomalies. Here, penalties are assigned in proportion to the physical differences that cause an anomaly, ensuring that the penalties are more reflective of real-world conditions:

$$p_{ij} = \frac{w_j - w_i}{\sum_{a=0}^{n-1} \sum_{b=i+1}^{n} (w_b - w_a)}. \quad (4)$$

In (4), $w_i$ represents the physical quantity corresponding to the severity level index $i$. For instance, in the dataset used herein [4], the physical quantities are the mass that causes unbalance and the defect length that causes misalignment. For bearing defects, the physical quantity is the defect length associated with the characteristic frequencies of ball pass frequency inner race (BPFI) and BPFO. The physics penalty, unlike the uniform or index penalties, requires precise physical quantities corresponding to the severity of the defects, which can make its application somewhat more challenging.

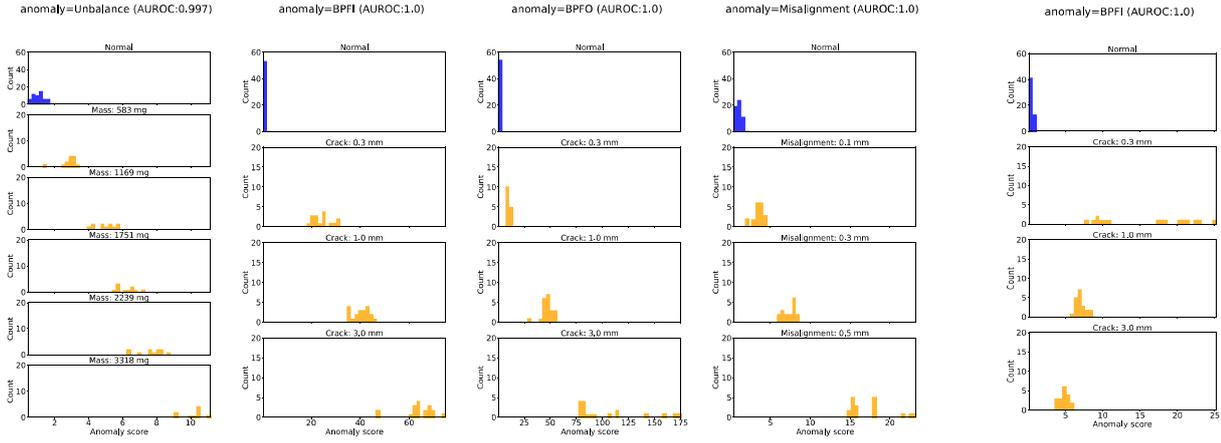

Fig. 1 Histogram of anomaly scores obtained using index 6 model (Rotation, BPFO, DL).

Fig. 2 Histogram of anomaly scores obtained using index 1 model. (BPFI)

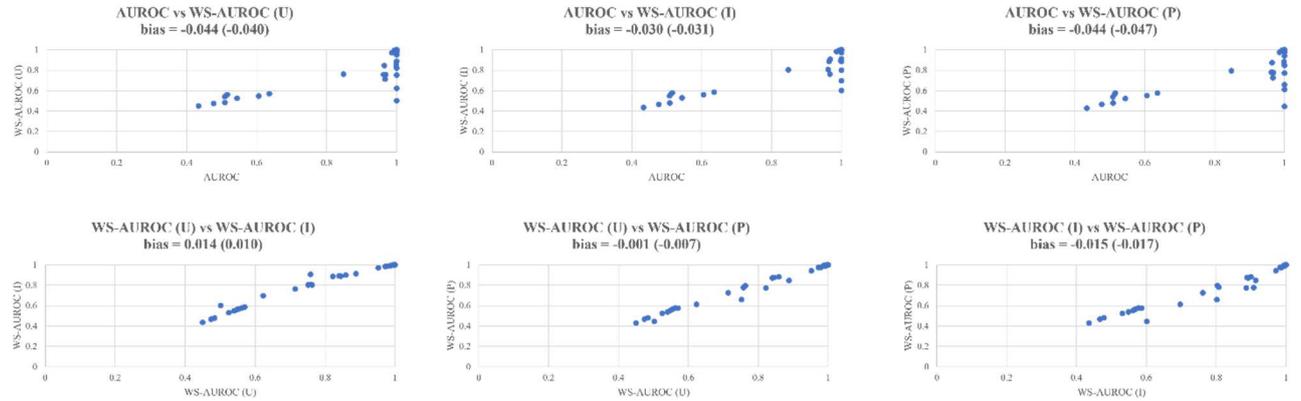

Fig. 3 Comparison of performance metrics. Bias means average of subtraction of x from y. The values outside the parentheses represent the bias of the entire dataset, and the values inside the parentheses represent the bias for the cases in which AUROC=1.

## III. PROPOSED ANOMALY DETECTION MODEL

### A. Frequency components of harmonics

The vibration dataset [4] simulates anomalies such as unbalance, misalignment, and bearing faults, which cause cracking of the inner and outer bearing races. For unbalance, the magnitude of the rotation frequency component $f_r$ increases compared to that under normal conditions. For misalignment, generally, the magnitudes of the $2f_r, 4f_r, \ldots$ components increase. For BPFI and BPFO, the magnitudes of the harmonics of their characteristic frequency increase. In the experiment conducted in [4], the rotation frequency was 50.17 Hz, BPFI characteristic frequency was 272.07 Hz, and BPFO characteristic frequency was 179.43 Hz. In the presence of these anomalies, we constructed vectors by using the magnitude and phase information of up to the fourth harmonic of $f_r$. Because we used four accelerometers, the length of each vector was 32. Additionally, we created separate vectors of length 32 for BPFI and BPFO by using their characteristic frequencies.

### B. Anomaly score obtained from KNN

We applied the KNN method with $k=1$ to calculate the minimum distance to the nearest neighbor. Then, we calculated the minimum distance of validation data and test data. Subsequently, we normalized the test-data distances by using the minimum and maximum validation-data distances.

### C. FFT and 1DCNN as feature extractor

We used FFT and 1DCNN to extract feature vectors and used KNN to score the test data. FFT components of up to 1 kHz were input into the 1DCNN, which distinguished torque loads (0 Nm, 2 Nm, 4 Nm) based on the data obtained using four accelerometers. To enhance feature representation learning, we used MixUp [6] with a uniform distribution. The model was trained over 20 epochs by employing categorical cross entropy loss, a learning rate of 1e-3, a batch size of 16, and the Adam optimizer [7]. The model with the lowest validation loss was used for anomaly detection. For the downstream task, we extracted feature vectors from the penultimate layer of the 1DCNN and trained a KNN-based model on the training-data feature vectors. For the validation and test datasets, we used the KNN model to calculate the minimum distances among the data obtained using four accelerometers, which yielded four anomaly scores per event. These scores were summed to obtain the final anomaly scores. The test-data anomaly scores were normalized using the minimum and maximum validation-data anomaly scores.

### D. Ensemble

By combining the four test anomaly scores, we created various combinations of anomaly detectors by using ensemble methods.

For convenience, we refer to the three anomaly scores obtained in sections III-A and III-B as Rotation, BPFI, and

BPFO, and the single anomaly score obtained in Section III-C as DL. The experimental results obtained by summing various combinations of these anomaly scores indicate that the summation of Rotation, BPFO, and DL yielded the best performance.

## IV. Experiment

The data used in the experiment consist of 432 normal samples for training, 54 normal samples for validation, and 270 samples for testing; the test samples contained 54 samples each for the normal, unbalance, BPFI, BPFO, and misalignment conditions. The unbalance anomaly samples had five severity levels (583 mg, 1169 mg, 1751 mg, 2239 mg, and 3318 mg); the BPFI and BPFO anomaly samples had three severity levels each (0.3 mm, 1.0 mm, and 3.0 mm); and the misalignment anomaly samples had three severity levels (0.1 mm, 0.3 mm, 0.5 mm). When the 1DCNN was not utilized, KNN, a non-parametric method, was run only once. However, when the 1DCNN was used, the experiment was conducted six times with different random seeds. AUROC and WS-AUROC were used as the performance metrics. By using 54 normal test samples and 54 anomaly test samples for each anomaly type, we calculated the AUROC and three types of WS-AUROC (Uniform, Index, and Physics) for four pairs of datasets, resulting in 16 individual performance metrics. Additionally, we calculated the averages of AUROC and three types of WS-AUROC.

## V. Result

Tables I–III summarize the performance metrics for different combinations of anomaly detection methods applied to various anomaly types. According to Table III, the combination of Rotation, BPFO, and DL yielded the highest average WS-AUROC anomaly score for the Uniform, Index, and Physics penalties, outperforming the other combinations. The anomaly score distributions of this combination are plotted in Fig. 1. The results show a clear distinction between the normal and anomaly conditions and separate the anomaly score distributions by severity levels.

According to Table I, the performance of BPFI at index 1 indicates that a high AUROC does not guarantee a high WS-AUROC. To achieve a high WS-AUROC, the model must distinguish between the normal and anomaly conditions and ensure that the anomaly score distribution is well ordered by severity levels. Fig. 2 illustrates this requirement, where the normal and anomaly conditions are well separated but the severity level anomaly score distributions are in reverse order, resulting in a lower WS-AUROC.

We plotted scatter plots and calculated biases for all results and cases with AUROC = 1, as shown in Fig. 3, to identify the most sensitive performance metric. A negative bias indicates higher sensitivity for the y-axis metric. The AUROC vs. WS-AUROC plots and bias calculations revealed WS-AUROC's sensitivity, as it did not consistently show high values near AUROC = 1. Among WS-AUROC metrics, the physics penalty was the most sensitive. Although the uniform penalty also showed sensitivity, it misaligned with the intended penalty philosophy by assigning constant penalties to low severity level differences. The physics penalty maintained this philosophy and proved to be the most sensitive.

## VI. Conclusion

We proposed WS-AUROC, which incorporates penalties based on severity level differences, to facilitate more refined evaluations than that possible with AUROC. In the experiments, the highest WS-AUROC score was obtained by using a combination of Rotation, BPFO, and DL anomaly scores. However, a high AUROC did not always correspond to a high WS-AUROC, highlighting the issues caused by reversed order of anomaly scores between different severity levels. We experimented with different penalty assignment methods for WS-AUROC, including the uniform, index, and physics-based penalties. The physics-based method demonstrated the highest sensitivity, adhering to the philosophy that greater severity level differences should incur higher penalties.

It should be noted that implementing WS-AUROC requires more detailed information than simpler metrics like AUROC, specifically regarding severity levels and associated penalties. This requirement can present challenges, especially when such data is not readily available, but it allows for a more precise and sensitive assessment of anomaly detection performance.

In future work, WS-AUROC can be applied to datasets beyond vibration, such as audio data [8], or even current data [1], [2]. Additionally, defining and utilizing other threshold-independent metrics, such as Partial Area Under the ROC Curve (pAUC), Area Under the Precision-Recall Curve (PR AUC), and F1-EV [9], can provide a more nuanced evaluation of models.

## Acknowledgment

This work was supported by the National Research Foundation of Korea(NRF) grant funded by the Korea government(MSIT) (No. RS-2024-00350917).